\documentclass[10pt,journal,compsoc]{IEEEtran}
\ifCLASSOPTIONcompsoc
  \usepackage[nocompress]{cite}
\else
  \usepackage{cite}
\fi
\ifCLASSINFOpdf
  \usepackage[pdftex]{graphicx}
\else
\fi
\usepackage{amsmath,amsfonts}
\usepackage{bm}

\usepackage{algorithm}
\usepackage{algorithmic}
\ifCLASSOPTIONcompsoc
 \usepackage[caption=false,font=footnotesize,labelfont=sf,textfont=sf]{subfig}
\else
 \usepackage[caption=false,font=footnotesize]{subfig}
\fi

\usepackage{color}
\usepackage[table]{xcolor}
\usepackage{multirow}
\usepackage{tikz}
\usepackage{pgfplots}
\pgfplotsset{compat=1.18}
\makeatletter
\newcommand{\thickhline}{%
	\noalign {\ifnum 0=`}\fi \hrule height 1pt
	\futurelet \reserved@a \@xhline
}
\makeatother

\definecolor{mygray}{gray}{.9}

\begin{document}
%
% paper title
% Titles are generally capitalized except for words such as a, an, and, as,
% at, but, by, for, in, nor, of, on, or, the, to and up, which are usually
% not capitalized unless they are the first or last word of the title.
% Linebreaks \\ can be used within to get better formatting as desired.
% Do not put math or special symbols in the title.
\title{V2X-PC: Vehicle-to-everything Collaborative Perception via Point Cluster}
%
%
% author names and IEEE memberships
% note positions of commas and nonbreaking spaces ( ~ ) LaTeX will not break
% a structure at a ~ so this keeps an author's name from being broken across
% two lines.
% use \thanks{} to gain access to the first footnote area
% a separate \thanks must be used for each paragraph as LaTeX2e's \thanks
% was not built to handle multiple paragraphs
%
%
%\IEEEcompsocitemizethanks is a special \thanks that produces the bulleted
% lists the Computer Society journals use for "first footnote" author
% affiliations. Use \IEEEcompsocthanksitem which works much like \item
% for each affiliation group. When not in compsoc mode,
% \IEEEcompsocitemizethanks becomes like \thanks and
% \IEEEcompsocthanksitem becomes a line break with idention. This
% facilitates dual compilation, although admittedly the differences in the
% desired content of \author between the different types of papers makes a
% one-size-fits-all approach a daunting prospect. For instance, compsoc 
% journal papers have the author affiliations above the "Manuscript
% received ..."  text while in non-compsoc journals this is reversed. Sigh.

\author{Si Liu, Zihan Ding, Jiahui Fu, Hongyu Li, Siheng Chen, Shifeng Zhang, Xu Zhou% <-this % stops a space
\IEEEcompsocitemizethanks{\IEEEcompsocthanksitem Si Liu, Zihan Ding, Jiahui Fu and Hongyu Li are with the Institute of Artificial Intelligence, Beihang University, Beijing 100191, China, and also with the Hangzhou Innovation Institute, Beihang University, Beijing 100191, China.%\protect\
% note need leading \protect in front of \\ to get a newline within \thanks as
% \\ is fragile and will error, could use \hfil\break instead.
% E-mail: see http://www.michaelshell.org/contact.html
\IEEEcompsocthanksitem Siheng Chen is with the Cooperative Medianet Innovation Center (CMIC), Shanghai Jiao Tong University, Shanghai, China, and also with Shanghai AI Laboratory, Shanghai, China.
\IEEEcompsocthanksitem Shifeng Zhang and Xu Zhou are with Sangfor Technologies Inc.% <-this % stops a space
%\thanks{Manuscript received April 19, 2005; revised August 26, 2015.}
}}

% note the % following the last \IEEEmembership and also \thanks - 
% these prevent an unwanted space from occurring between the last author name
% and the end of the author line. i.e., if you had this:
% 
% \author{....lastname \thanks{...} \thanks{...} }
%                     ^------------^------------^----Do not want these spaces!
%
% a space would be appended to the last name and could cause every name on that
% line to be shifted left slightly. This is one of those "LaTeX things". For
% instance, "\textbf{A} \textbf{B}" will typeset as "A B" not "AB". To get
% "AB" then you have to do: "\textbf{A}\textbf{B}"
% \thanks is no different in this regard, so shield the last } of each \thanks
% that ends a line with a % and do not let a space in before the next \thanks.
% Spaces after \IEEEmembership other than the last one are OK (and needed) as
% you are supposed to have spaces between the names. For what it is worth,
% this is a minor point as most people would not even notice if the said evil
% space somehow managed to creep in.

% The paper headers
\markboth{IEEE TRANSACTIONS ON PATTERN ANALYSIS AND MACHINE INTELLIGENCE}{}
\IEEEtitleabstractindextext{%
\begin{abstract}
The objective of the collaborative vehicle-to-everything perception task is to enhance the individual vehicle's perception capability through message communication among neighboring traffic agents.
Previous methods focus on achieving optimal performance within bandwidth limitations and typically adopt BEV maps as the basic collaborative message units.
However, we demonstrate that collaboration with dense representations is plagued by object feature destruction during message packing, inefficient message aggregation for long-range collaboration, and implicit structure representation communication.
To tackle these issues, we introduce a brand new message unit, namely point cluster, designed to represent the scene sparsely with a combination of low-level structure information and high-level semantic information.
The point cluster inherently preserves object information while packing messages, with weak relevance to the collaboration range, and supports explicit structure modeling.
Building upon this representation, we propose a novel framework V2X-PC for collaborative perception.
This framework includes a Point Cluster Packing (PCP) module to keep object feature and manage bandwidth through the manipulation of cluster point numbers.
As for effective message aggregation, we propose a Point Cluster Aggregation (PCA) module to match and merge point clusters associated with the same object.
To further handle time latency and pose errors encountered in real-world scenarios, we propose parameter-free solutions that can adapt to different noisy levels without finetuning.
Experiments on two widely recognized collaborative perception benchmarks showcase the superior performance of our method compared to the previous state-of-the-art approaches relying on BEV maps.
\end{abstract}

% Note that keywords are not normally used for peerreview papers.
}

% make the title area
\maketitle

% To allow for easy dual compilation without having to reenter the
% abstract/keywords data, the \IEEEtitleabstractindextext text will
% not be used in maketitle, but will appear (i.e., to be "transported")
% here as \IEEEdisplaynontitleabstractindextext when compsoc mode
% is not selected <OR> if conference mode is selected - because compsoc
% conference papers position the abstract like regular (non-compsoc)
% papers do!
\IEEEdisplaynontitleabstractindextext
% \IEEEdisplaynontitleabstractindextext has no effect when using
% compsoc under a non-conference mode.

% For peer review papers, you can put extra information on the cover
% page as needed:
% \ifCLASSOPTIONpeerreview
% \begin{center} \bfseries EDICS Category: 3-BBND \end{center}
% \fi
%
% For peerreview papers, this IEEEtran command inserts a page break and
% creates the second title. It will be ignored for other modes.
\IEEEpeerreviewmaketitle

\ifCLASSOPTIONcompsoc
\IEEEraisesectionheading{\section{Introduction}\label{sec:introduction}}
\else
\section{Introduction}
\label{sec:introduction}
\fi

\begin{figure*}[!t]
\centering
\includegraphics[width=\textwidth]{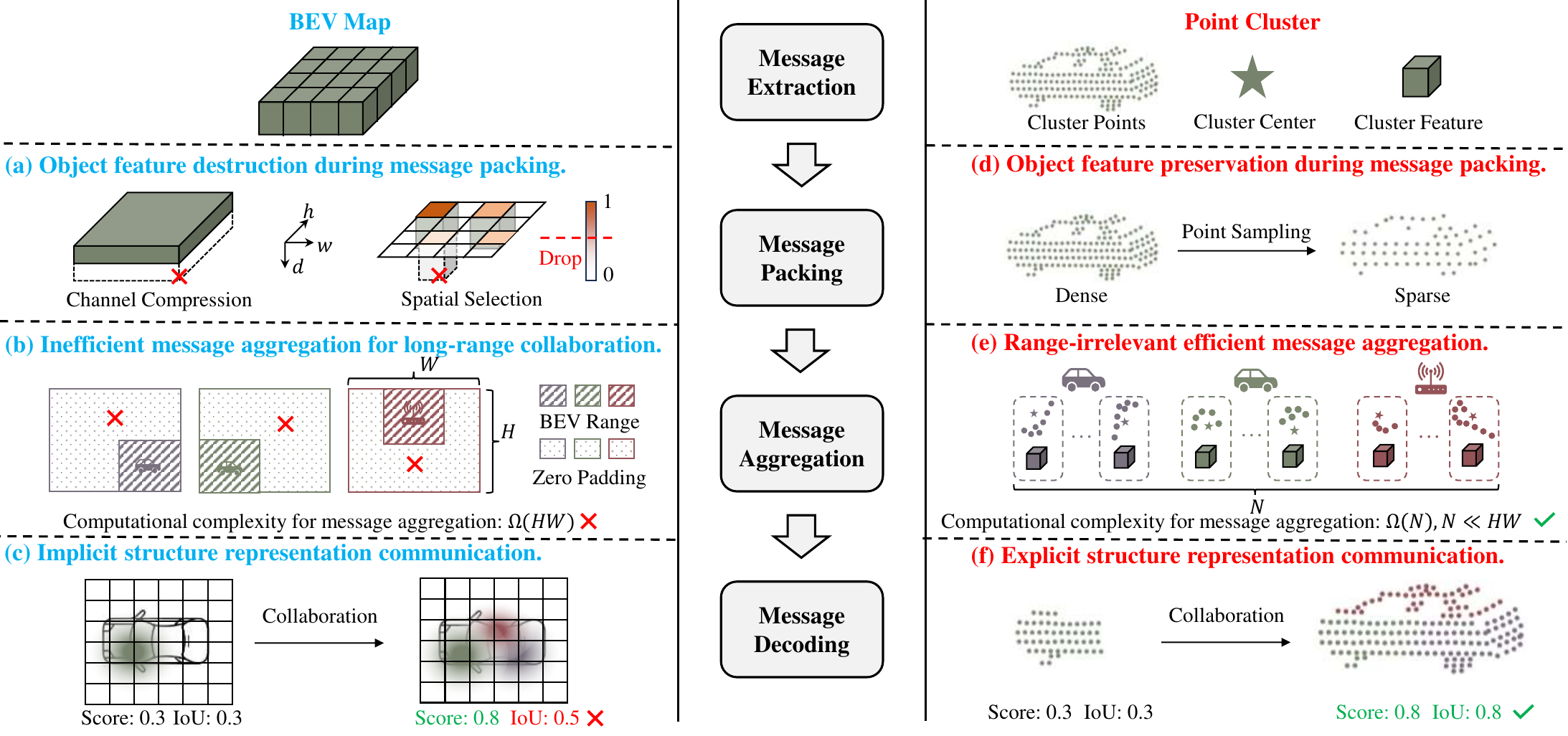}
\caption{Comparison of using BEV map and point cluster as the basic interaction unit for collaborative perception.
(Left) Channel compression and spatial selection applied on BEV maps for reducing bandwidth usage suffer from feature degradation and object loss, respectively;
The aggregation process of BEV maps introduces unnecessary zero-padding and the computational complexity increases quadratically with the expansion of the communication range;
Voxelization-induced implicit structure representation communication can lead to inaccurate prediction of box boundaries.
(Right) We can control the bandwidth usage of point clusters by sampling important points instead of compressing the high-level cluster feature;
Message aggregation based on point cluster only related to the number of potential objects in the scene;
We can complete the spatial structure of objects with the low-level point coordinates for predictions with high precision.
}
\label{fig:intro:motivation}
\end{figure*}

\IEEEPARstart{P}{erception} plays a pivotal role in autonomous driving as a crucial component, with its primary objective being to comprehend the surrounding environment and extract pertinent information to facilitate subsequent prediction and planning tasks.
With ongoing advancements in deep learning, individual perception has demonstrated remarkable progress in various tasks, such as detection~\cite{he2017mask, shi2020pv, zhou2018voxelnet}, segmentation~\cite{cheng2022masked, qi2017pointnet++, pan2023baeformer}, and tracking~\cite{chen2021transformer, yin2021center, zhou2022pttr}.
While showing potential, this approach often encounters problems such as occlusion, which arises from the restricted sight-of-view of individual agents, and faces safety challenges.
To overcome this bottleneck issue, collaborative perception in vehicle-to-everything (V2X) autonomous driving has captured the attention of both academic and industrial sectors over the past few years.
The objective is to enhance the perception capabilities of individual vehicles by utilizing complementary information exchanged between surrounding traffic agents, such as vehicles and infrastructure.

In order to achieve collaborative perception, recent efforts have made valuable contributions in terms of high-quality real and simulated datasets~\cite{yu2022dair, xu2022opv2v, li2022v2x, xu2022v2x, xu2023v2v4real}, as well as effective solutions~\cite{xu2023cobevt, li2021learning, wang2020v2vnet, xu2022v2x, hu2022where2comm, chen2019f, xu2022opv2v}.
As shown in the middle of Figure~\ref{fig:intro:motivation}, a prevalent approach in this field involves four stages, i.e., message extraction, message packing, message aggregation, and message decoding.
Initially, each agent transforms the scene point clouds scanned by itself into intermediate representations, serving as the basic collaborative message units. 
In the process of message packing, each agent acts as a sender to strategically pack self-aware informative units and transmit compressed features to the ego agent as a receiver under the restrictions of transmission bandwidth.
Upon receiving messages from other agents, the ego agent utilizes diverse multi-source aggregation techniques to enhance its scene representations for a holistic perception, based on which we can decode the final perception results. 
The main challenge in this field is to efficiently leverage limited bandwidth to attain optimal performance.
This includes how other agents can maximize compression during message packing while retaining useful information largely intact, and how the ego agent can leverage them to reconstruct and merge into a unified scene representation during message aggregation.

Despite significant development, we argue that existing collaborative perception methods employing dense BEV maps as basic collaborative message units still have several limitations:
1) \textit{Object feature destruction during message packing} (Figure~\ref{fig:intro:motivation} (a)). 
Existing BEV-based methods often use channel compression~\cite{xu2022v2x} or spatial selection~\cite{hu2022where2comm,wang2023umc} to balance perception performance and communication bandwidth during message packing. 
Channel compression enables the sender to preserve spatially complete scene information, while the receiver may suffer from object feature degradation during reconstruction due to heterogeneity representation across channels.
In contrast, spatial selection transmits only informative regions pointed out by spatial confidence maps, which may result in potential object loss when bandwidth constraints become more stringent.
2) \textit{Inefficient message aggregation for long-range collaboration} (Figure~\ref{fig:intro:motivation} (b)).
Collaboration can bring a larger perception range to the ego agent, but computational complexity also grows quadratically with the expansion of dense BEV feature maps. 
Moreover, limited by the square map structure and convolution operation requirements, the received BEV features are inevitably filled to the same shape for aggregation with zero paddings. However, calculating the non-overlapped areas between agents area is essentially unnecessary.
3) \textit{Implicit structure representation communication} (Figure~\ref{fig:intro:motivation} (c)).
The voxelization operation sacrifices 3D geometric details in comparison to the raw point clouds.
While aggregating BEV representations from different agents can enhance the response of potential object regions, the precision of predicted box boundaries may be constrained by incomplete object structure modeling.

Considering the above issues, a question naturally arises: \textit{Is there a better intermediate representation that can serve as the collaborative message unit to unleash the performance of collaborative perception}?
Recently, sparse detectors~\cite{shi2019pointrcnn,yan2018second,fan2022fully} have demonstrated significant advancements in LiDAR-based 3D object detection, which directly operate on 3D point clouds and fundamentally avoid the information loss caused by voxel quantization.
In light of this, we propose a new form of collaborative message unit, called point cluster, which consists of point coordinates representing the object structure, a cluster center representing the object position, and a cluster feature representing the high-level semantics of the object.
Thus, the point cluster can describe information from point clouds comprising an object instance.
The point cluster has merits in three aspects as the collaborative message unit compared to the BEV representation:
1) \textit{Object feature preservation during message packing.} (Figure~\ref{fig:intro:motivation} (d)). 
Point clusters inherently contain only the information of foreground objects present in the scene, eliminating the need for filtering out irrelevant backgrounds by handcrafted rules. 
Moreover, we can control the transmission bandwidth by explicitly reducing the number of points, as opposed to implicitly compressing feature channels.
2) \textit{Range-irrelevant efficient message aggregation} (Figure~\ref{fig:intro:motivation} (e)). 
The number of point clusters is more related to the number of objects in the scene rather than the collaboration range. 
Furthermore, point clusters can be conveniently associated with the same object and aggregated through set merging, without the need for padding to the same shape of joint field of view.
3) \textit{Explicit structure representation communication} ((Figure~\ref{fig:intro:motivation} (f))). 
Point clusters fully preserve the geometric structural information of objects in the original coordinate space, enabling fine-grained point alignment and complementary structure information fusion between different agents, which can improve the precision of predictions.

With point cluster as the collaborative message unit, we propose a novel collaborative perception framework in this paper, namely V2X-PC, for V2X autonomous driving.
We utilized a Point Cluster Encoder (PCE) to encode the original point clouds of all agents into corresponding point cluster representations for message extraction.
During message packing, every agent distributes its perceived set of point clusters and localization position to surrounding agents.
To cope with different bandwidth constraints, we propose a Point Cluster Packing (PCP) module to flexibly control the number of points contained in each point cluster while maintaining their geometric structure as much as possible.
Compared with BEV-based methods, our PCP only operates on low-level object representation and avoids high-level object information loss, thereby maintaining relatively high precision and recall under strict bandwidth constraints.
After receiving messages from other agents, we propose a Point Cluster Aggregation (PCA) module to integrate point clusters from other agents to achieve a comprehensive understanding of the surrounding scene for the ego agent.
In detail, we match point clusters from different agents according to the distance between their cluster centers.
After, we merge point clusters identified to the same object to a new point cluster, based on which we output the final predictions.
The computational complexity of our PCA module is only related to the number of point clusters and there is no padding operation, which is more effective and efficient for long-range collaboration than BEV-based approaches.

Besides the bandwidth constraints, pose error~\cite{lu2023robust} and time latency~\cite{lei2022latency,wei2024asynchrony} are two common challenges that can affect the robustness of collaborative perception methods.
To explicitly solve the pose error, we propose to reformulate cluster centers as vertices of a graph, on which we can apply the traditional graph optimization algorithm to promote pose consistency between agents and point clusters.
In order to compensate for the time latency, we link point clusters of the same object along the time dimension and measure its speed for position prediction in the current timestamp.
Our graph optimization and speed estimation methods possess a strong generalization capability as they do not rely on any training parameters during the optimization process. 
This enables our method to adapt to varying levels of pose errors and time latencies with ease.

To validate the effectiveness of our proposed V2X-PC, we conduct experiments on two popular V2X benchmarks, V2X-Set~\cite{xu2022v2x} and DAIR-V2X~\cite{yu2022dair}, respectively. 
Our method achieves the new state-of-the-art performances compared to previous BEV-based collaborative perception methods.
To further demonstrate the benefits brought by collaborative perception to single-vehicle perception, we divided the target objects into different groups according to the number of their occupied points observed by the ego agent and other agents.
The less the object is observed by the ego agent, the more the object needs to be collaboratively perceived.

In summary, this paper has the following contributions:
\begin{itemize}
    \item We propose a brand new collaborative message unit called point cluster, based on which we introduce an effective, efficient, and robust V2X collaborative perception framework called V2X-PC.
    \item To achieve bandwidth-friendly message packing, we propose a Point Cluster Packing (PCP) module that controls bandwidth usage by adjusting the point number in point clusters, which avoids high-level object information loss.
    \item During message aggregation, we propose a Point Cluster Aggregation (PCA) module to complete object information by merging point clusters, whose computational complexity is only related to the object number in the scene.
    It makes our method friend to long-range collaborative perception.
    \item As for robustness, we propose parameter-free solutions for the pose error and time latency problems based on the low-level object information contained in the point clusters, which shows great generalization capability.
    \item We propose a new metric for collaborative perception, which can evaluate the benefits of collaborative perception in a more intuitive way.
    Extensive experiments on two popular collaborative perception benchmarks validate that our methods outperform previous BEV-based methods.
    We will release the code for future academic research.
\end{itemize}

\section{Related Work}

\subsection{Collaborative Perception}

Collaborative perception can be systematically classified into three primary types based on distinct fusion stages: early, intermediate, and late collaboration.
Early collaboration~\cite{chen2019cooper, arnold2020cooperative} integrates raw sensory data from vehicles and infrastructure, providing a comprehensive perspective. 
For example, Arnold \textit{et al}.~\cite{arnold2020cooperative} first preprocess respective point clouds onboard each sensor in the global coordinate system.
Subsequently, these point clouds are transmitted to a centralized fusion system, where they are combined into a holistic point cloud and inputted into the detection model.
However, this type of collaboration can quickly overwhelm the communication network with substantial data traffic, making it impractical for most purposes.
In contrast, late collaboration~\cite{shi2022vips, zeng2020dsdnet, rauch2012car2x, glaser2023we} conducts the collaboration in the output space, which is bandwidth-economic but sensitive to noise and estimation errors.
Intermediate collaboration enables the exchange of intermediate features created by the involved agents and has demonstrated significant potential in recent years.
Considering the bandwidth constraints in practical application scenarios, various cooperation strategies are proposed to decide who~\cite{liu2020who2com}, when~\cite{liu2020when2com}, and where~\cite{hu2022where2comm, wang2023umc} to communicate. 
After receiving features from other agents, existing methods adopt attention mechanism~\cite{xu2022opv2v, xu2022v2x, xu2023cobevt, zhang2022multi, wang2023core, yang2023spatio}, graph neural network~\cite{wang2020v2vnet, li2021learning}, maxout~\cite{bai2022pillargrid, guo2021coff, qiao2023adaptive}, and addition~\cite{marvasti2020cooperative} to aggregate complementary scene information.
Different from previous intermediate collaboration methods adopting BEV maps as the collaborative message units, the introduced V2X-PC in this paper is based on our newly proposed point cluster and achieves state-of-the-art performance.
\begin{figure*}[!t]
\centering
\includegraphics[width=\textwidth]{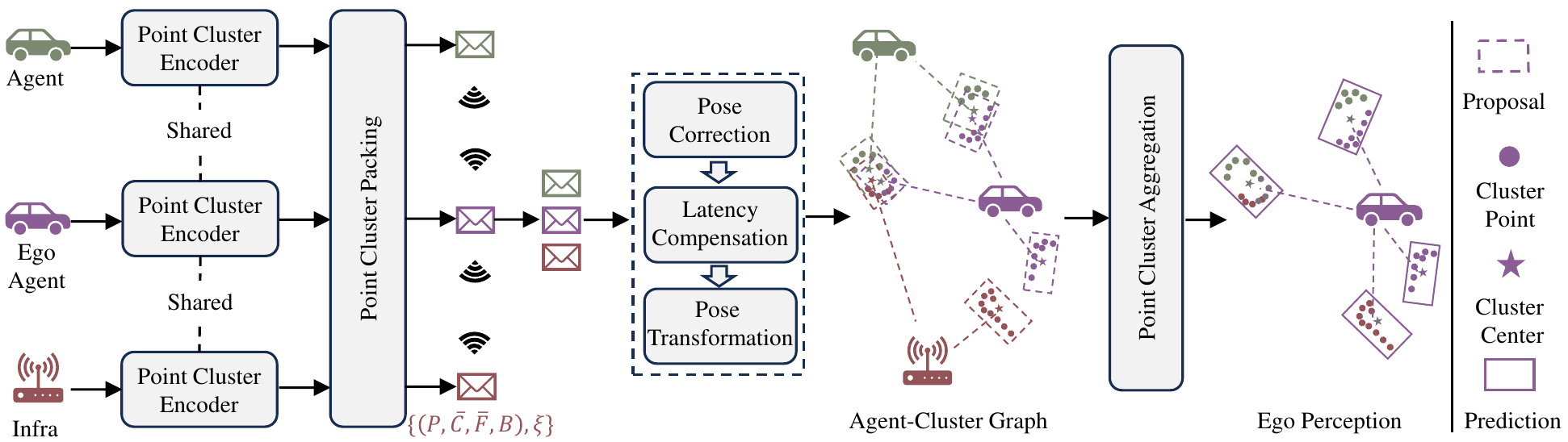}
\caption{Overall architecture of our method. Point Cluster Encoder (PCE) extracts point cluster representations of each agent's observation. Then a Point Cluster Packing (PCP) module is applied to filter noisy point clusters and correct the point coordinates in those kept. We can control the bandwidth usage by reducing the number of points involved in point clusters. After receiving messages from other agents, we address the time latency and pose error problems, and transform all point clusters to the coordinate space of the ego agent. Finally, we use a Point Cluster Aggregation (PCA) module to complete object information and output the predictions.}
\label{fig:method:framework}
\end{figure*}

\subsection{Sparse Detectors}

Mainstream dense 3D detectors typically convert orderless point clouds into intermediate compact representations, which include images from BEV or frontal view~\cite{lang2019pointpillars,yang2018pixor}, as well as voxels that are equally distributed~\cite{zhou2018voxelnet,yan2018second}.
However, this process inevitably sacrifices geometric details and introduces quantization errors, resulting in performance bottlenecks.
To address this issue, point-based detectors have emerged as a popular research topic.
PointRCNN~\cite{shi2019pointrcnn} is recognized as groundbreaking research in the advancement of this line of work.
Building on this approach, 3DSSD~\cite{yang20203dssd} eliminates the feature propagation layers and the refinement module to accelerate point-based methods.
Taking inspiration from Hough voting, VoteNet~\cite{qi2019deep} initially casts votes for object centroids and subsequently generates high-quality proposals based on the voted center.
IA-SSD~\cite{zhang2022not} prioritizes foreground information and automatically selects the crucial instance points that are sparse in number.
Despite multiple attempts to speed up the point-based method, the time-consuming neighborhood query~\cite{qi2017pointnet++} remains impractical for large-scale point clouds that have over 100k points.
FSD~\cite{fan2022fully} is the pioneering fully sparse 3D object detector, which treats instances as groups and gets rid of the dependence on the neighborhood query.
VoxelNeXt~\cite{chen2023voxelnext} simplifies the fully sparse architecture even further by utilizing only voxel-based designs.
FlatFormer~\cite{liu2023flatformer} substitutes the 3D backbone in FSD with a novel transformer-based sparse backbone to reduce latency.
In this paper, we study the shortcomings of BEV-based collaboration methods and propose the point cluster as the collaborative message unit, which keeps the low-level structure information to support effective and efficient collaboration.

\section{Method}

To begin with, we will present a concise introduction to the overall problem formulation of collaborative perception in \ref{sec:problem_formulation}. 
Next, we will present an overview of our V2X-PC framework in \ref{sec:overview_of_framework}, followed by an introduction to the extraction process of our proposed point cluster in \ref{sec:point_cluster_extraction}. 
The point cluster packing and aggregation processes are described in \ref{sec:point_cluster_distribution} and \ref{sec:point_cluster_aggregation}, respectively. 
Finally, we will address time latency and pose error problems to improve our method's robustness in~\ref{sec:robustness}.

\subsection{Problem Formulation}
\label{sec:problem_formulation}

Suppose there are $N_\text{agent}$ agents present in the scene, whose observations and the ground truth annotation are denoted as $\{\mathcal{X}^{i}\}_{i=1}^N$ and $\{\mathcal{Y}^{i}\}_{i=1}^N$, respectively. 
Collaborative perception aims to maximize the perception performance of all agents while taking into account the constraint on available bandwidth $\beta$, which can be formulated as:
\begin{equation}
    \mathop{\arg\max}_{\theta, \mathcal{M}} \sum_{i=1}^N g(\Phi_\theta(\mathcal{X}^{i;t}, \{\mathcal{M}^{j;t-\tau^{{j\rightarrow i};t}}\}_{j=1,j\neq i}^N), \mathcal{Y}^{i;t}), \\
\end{equation}
\begin{equation}
    \text{subject to} \sum_{i=1}^N\sum_{j=1,j\neq i}^N|\mathcal{M}^{j;t-\tau^{j\rightarrow i;t}}|\leq \beta.
\end{equation}
Here, $\theta$ denotes trainable parameters of the network $\Phi$ and $g(\cdot, \cdot)$ is the evaluation metric. 
$\mathcal{M}^{j;t-\tau^{{j\rightarrow i};t}}$ is the message transmitted from the $j$-th agent at the timestamp $t-\tau^{{j\rightarrow i};t}$ and received by the $i$-th agent at the timestamp $t$, where $\tau^{{j\rightarrow i};t}$ is the transmission latency. 
It includes the $j$-th agent's extracted collaborative message units $\bm{M}^j$ and 6DoF pose $\xi^j$.
Note that 
1) there is no collaboration when $\beta=0$ and the objective described above reflects the single-agent perception performance;
2) when accounting for pose error, it is necessary to correct the pose $\xi^j$ for feature alignment;
3) we will omit the superscripts $t$ related to the timestamp in the following when $\tau=0$ for simplicity.

\subsection{Overview of V2X-PC}
\label{sec:overview_of_framework}

Figure~\ref{fig:method:framework} illustrates the overall architecture of our proposed method.
Initially, the raw point clouds of all agents are processed by a shared Point Cluster Encoder (PCE), which segments foreground points on the surface of objects and divides them into clusters based on the distance metric.
From each of these point clusters, we extract and formulate the point coordinates, the center coordinates, and the cluster feature as corresponding intermediate representations.
Then we propose a Point Cluster Packing (PCP) module to filter noisy background clusters and correct the involved points of foreground clusters via proposal generation. 
The reduction in bandwidth usage of the point cluster can be achieved by decreasing the number of included points. 
After receiving messages from other agents, we address the pose error and time latency with parameter-free approaches and then align the coordinate space of point clusters from multiple agents via pose transformation, which allows the ego agent to obtain an agent-cluster graph as the comprehensive scene representation from its own coordinate space.
As for message aggregation, we propose a Point Cluster Aggregation (PCA) module, where point cluster matching is performed to find point clusters belonging to the same object and merge them into a new point cluster that contains complete object information with linear complexity.
Finally, we refine cluster features of point clusters via point-based operators, based on which we output the final detection results.

\subsection{Point Cluster Encoder}
\label{sec:point_cluster_extraction}

Since the original BEV representations for collaborative perception suffer from high-level object information loss, inefficient message aggregation for long-range collaboration, and implicit structure representation communication, we propose a brand new representation called point cluster to address these issues.
For simplicity, we omit the superscript $i$ and $j$ in this section because all agents share the same point cluster extraction pipeline.
We adopt the encoder-decoder point cloud backbone~\cite{shi2020points} based on 3D sparse convolution and deconvolution as the sparse voxel feature extractor.
To construct each point feature, we concatenate the voxel feature where the point is located and the corresponding offset from the point to the voxel center.
These point features are then passed through a multi-layer perceptron (MLP) for foreground segmentation.
Since objects are naturally well-separated, 3D box annotations in autonomous driving scenes directly provide semantic masks for supervision~\cite{shi2019pointrcnn}.
We use focal loss~\cite{lin2017focal} for segmentation loss, denoted as $\mathcal{L}_{\text{seg}}$.
For foreground points, we use another MLP to predict their offsets to the corresponding object centers, which is supervised by L1 loss~\cite{ren2015faster}, denoted as $\mathcal{L}_{\text{center}}$.
Next, we measure the distance among the predicted centers of foreground points, where two points belong to the same point cluster if their predicted centers' Euclidean distance is smaller than a certain threshold $\epsilon_\text{point}$.

After we can directly extract cluster features via the combination of three basic components of point-based operators summarized by Fan \textit{et al},~\cite{fan2022fully}, i.e., \textit{cluster center}, \textit{pair-wise feature}, and \textit{cluster feature aggregation}. 
The cluster center is the representative point of a cluster, while the pair-wise feature specifies the pairing of cluster centers and their corresponding points for cluster-aware point feature extraction. 
Cluster feature aggregation involves using a pooling function to combine the point features of a cluster. 
Based on these elements, variants of point-based operators can be developed, such as PointNet~\cite{qi2017pointnet}, DGCNN~\cite{wang2019dynamic}, Meta-Kernel~\cite{fan2021rangedet}, and SIR~\cite{fan2022fully}.
We select SIR in this paper and stack $L_1$ layers to encode all cluster features in parallel.
To express more clearly, we take the processing process of the $q$-th cluster in the $l$-th SIR layer as an example.
Concretely, assume there are $N^q_\text{point}$ foreground points, we denote the included point coordinates and features as $\bm{P}^q\in\mathbb{R}^{N^q_\text{point}\times 3}$ and $\bm{F}_\text{point}^{q;l}\in \mathbb{R}^{N^q_\text{point}\times D}$, respectively, where $D$ is number of feature channels.
We take the average coordinates of all predicted cluster centers as the cluster center, denoted as $\bm{C}^q\in\mathbb{R}^{1\times 3}$.
The processing process of SIR can be formulated as:
\begin{equation}
    \widetilde{\bm{F}}_\text{point}^{q;l}=\text{MLP}([\bm{F}_\text{point}^{q;l};\bm{P}^q\ominus\bm{C}^q]),
\end{equation}
\begin{equation}
    \bm{F}_\text{point}^{k;l+1}=\text{MLP}([\widetilde{\bm{F}}_\text{point}^{q;l};\text{maxpool}(\widetilde{\bm{F}}_\text{point}^{q;l})]),
\end{equation}
where $[;]$ denotes concatenation along the channel dimension, $\ominus$ means applying subtraction on each point in $\bm{P}^q$, and $\bm{F}_\text{point}^{q;l+1}\in \mathbb{R}^{N_\text{point}^q\times D}$ is the processed point cluster feature.
We concatenate $\{\bm{F}_\text{point}^{q;l}\}_{l=1}^{L_1}$ along the channel dimension, and apply linear transformation and max-pooling on it to obtain the final cluster feature $\bm{F}^q\in \mathbb{R}^{1\times D}$.

\subsection{Point Cluster Packing}
\label{sec:point_cluster_distribution}

The intermediate representations of point clusters to be packed by the $i$-th agent can be formulated as:
\begin{equation}
    \bm{M}^i=\{\bm{m}^{i;q}\}_{q=1}^{N_\text{cluster}^i}=\{(\bm{P}^{i;q}, \bm{C}^{i;q}, \bm{F}^{i;q})\}_{k=1}^{N_\text{cluster}^i},
\end{equation}
where $\bm{m}^{i;q}$ is the representation of the $q$-th cluster and $N_\text{cluster}^i$ is the number of point clusters extracted by the $i$-th agent.
However, during the point cluster grouping process, errors may occur, resulting in the loss of points on the object's surface and the inclusion of background distractors.
In order to correct this issue, we propose to generate a proposal bounding box for each cluster. 
To achieve this, we feed the cluster features to two separate MLPs for proposal classification and regression. 
During the training phase, clusters are classified as positive if their predicted centers are located in the ground truth bounding boxes.
We adopt L1 loss~\cite{ren2015faster} and focal loss~\cite{lin2017focal} as regression loss $\mathcal{L}_{\text{reg}}$ and classification loss $\mathcal{L}_{\text{cls}}$, respectively.
After proposal generation, we only retain point clusters with positive proposals. 
The point coordinates in these clusters are overridden with the coordinates of points within the positive proposals.
Consider there are $N^i_\text{cluster+}$ positive clusters, we modify the formulation of $\bm{M}_i$ as following:
\begin{equation}
    \bm{M}^i=\{\bm{m}^{i;q}\}_{q=1}^{N_\text{cluster+}^i}=\{(\bm{P}^{i;q}, \bm{C}^{i;q}, \bm{F}^{i;q},\bm{B}^{i;q})\}_{q=1}^{N_\text{cluster+}^i}, 
\end{equation}
where $\bm{B}^{i;q}=(\hat{\bm{x}}, \hat{\bm{y}}, \hat{\bm{z}}, \hat{\bm{h}}, \hat{\bm{w}}, \hat{\bm{l}}, \bm{\alpha}, \hat{\bm{c}})$ includes the center coordinates, the size, the yaw angle and the confidence score of the proposal bounding box of the $q$-th cluster.
The final message distributed can be formulated as follows:
\begin{equation}
    \mathcal{M}^i=(\bm{M}^i,\xi^i),
\end{equation}
where $\xi^i$ is the 6DoF pose used for pose transforming in the later aggregation phase.

Optimizing the trade-off between perception performance and communication bandwidth is vital in collaborative perception.
Unlike BEV maps, our point clusters are inherently sparse in the spatial dimensions. Therefore, the bandwidth consumption in our method is mainly on the point coordinates (\emph{i.e.}~thousands of $3$-dimensional coordinates) rather than the object features (\emph{i.e.}~one $128$-dimensional features). Considering that the geometric structure of the object can be represented by keypoints, we adopt a sampling method for the object points to compress the transmission data.
As a result, the high-level object information can be kept during message packing, which is more beneficial for tasks considering semantic information like 3D detection.

To reduce the redundancy of points while preserving valuable information as much as possible, we propose a Semtanic and Distribution guided Farthest Point Sampling~(SD-FPS) method to obtain object keypoints, which effectively preserves the semantic and structural information of objects. 
Specifically, for points of each object, we adopt a selection approach similar to Farthest Point Sampling~(FPS) to obtain a sparse result. In particular, in addition to considering the distance between points, we also select keypoints based on the object's semantic confidence score and distribution density score.
The semantic confidence score is derived from the prediction results of the previous stage segmentation, with a higher score indicating richer semantic information. The distribution density score is obtained using kernel density estimation (KDE), where a higher score indicates sparser local points.

We show the details of applying SD-FPS on the $q$-th cluster of the $i$-th agent as an example in Algorithm~\ref{alg:point_sampling} and omit the superscripts for simplicity.
During each round of selection, we select the points that are more representative of the object's semantics and structure in the point cluster based on the measurement as:
\begin{equation}
    \widetilde{\bm{d}}_\text{point}=(\bm{s}_\text{f})^{\lambda_\text{s}} \cdot (\bm{s}_\text{d})^{\lambda_\text{d}} \cdot \bm{d}_\text{point},
\end{equation}
where $\bm{s}_\text{f}$ is the semantic score obtained from the segmentation module in PCE, $\bm{s}_\text{d}$ is the distribution score represented by the kernel density~\cite{hu2022point}, and $\bm{d}_\text{point}$ is the closest distance to the already selected key points.
$\lambda_\text{s}$ and $\lambda_\text{d}$ are the coefficients representing the importance of the semantic and distribution information, respectively.
The SD-FPS algorithm will reduce to S-FPS when $\lambda_\text{d}=0$ and D-FPS when $\lambda_\text{s}=0$, as well as vanilla FPS when both $\lambda_\text{d}=0$ and $\lambda_\text{s}=0$.

\begin{algorithm}[t]
    \caption{Semantic- and Distribution-guided Farthest Point Sampling Algorithm. $N_\text{point}$ is the number of input points and $N_\text{sample}=N_\text{point}\times \zeta$ is the number of sampled points controlled by a predefined sampling rate $\zeta$.}
    \label{alg:point_sampling}
    \begin{tabular}{l@{}l}
        \textbf{Input: } & coordinates $\bm{P}=\{\bm{p}^1,\dots,\bm{p}^{N_\text{fg}}\}\in \mathbb{R}^{N_\text{point}\times3}$;\\
        & semantic scores $\bm{S}_\text{f}=\{\bm{s}_\text{f}^1,\dots,\bm{s}_\text{f}^{N_\text{point}}\}\in \mathbb{R}^{N_\text{point}}$;\\
        & distribution scores $\bm{S}_\text{d}=\{\bm{s}_\text{d}^1,\dots,\bm{s}_\text{d}^{N_\text{point}}\}\in \mathbb{R}^{N_\text{point}}$.
    \end{tabular}
    
    \begin{tabular}{l@{}l}
        \textbf{Output: } & sampled key point set $\widetilde{\bm{P}}=\{\widetilde{\bm{p}}^1,\dots,\widetilde{\bm{p}}^{N_\text{sample}}\}$
    \end{tabular}
    
    \begin{algorithmic}[1]
        \STATE initialize an empty sampling point set $\widetilde{\bm{P}}$
        \STATE initialize a distance array $\bm{D}_\text{point}$ of length $N_\text{point}$ with all $+\infty$
        \STATE initialize a visit array $\bm{V}$ of length $N_\text{point}$ with all zeros
        \FOR{$n=1$ \TO $N_\text{sample}$}
            \IF{$n=1$}
                \STATE $o = \arg\max(\{\bm{s}_\text{f}^k+\bm{s}_\text{d}^k|\bm{V}^k=0\}_{k=1}^{N_\text{point}})$
            \ELSE
                \STATE $\widetilde{\bm{D}}_\text{point} = \{(\bm{s}_\text{f}^k)^{\lambda_\text{s}} \cdot (\bm{s}_\text{d}^k)^{\lambda_\text{d}} \cdot \bm{d}_\text{point}^k | \bm{V}^k = 0 \}_{k=1}^{N_\text{point}}$
                \STATE $o = \arg\max(\widetilde{\bm{D}}_\text{point})$
            \ENDIF
            \STATE add $\bm{P}^o$ to $\widetilde{\bm{P}}$, $\bm{V}^o = 1$
            \FOR{$u=1$ \TO $N$}
                \STATE $\bm{d}_\text{point}^u = \min(\bm{d}_\text{point}^u, \|\bm{p}^u - \bm{p}^o\|)$
            \ENDFOR
        \ENDFOR
        \RETURN $\widetilde{\bm{P}}$
    \end{algorithmic}
\end{algorithm}

\subsection{Point Cluster Aggregation}
\label{sec:point_cluster_aggregation}

After message communication, we need to appropriately aggregate point clusters from surrounding agents to form a holistic perception.
We take the aggregation process from the $j$-th agent to the $i$-th agent as an example, which can be extended to all agents easily.
Firstly, we align the coordinate space of $\bm{M}^j$ to that of $\bm{M}^i$ through the transform matrix calculated from $\xi^i$ and $\xi^j$.
The transformed message units from the $j$-th agent are denoted as $\bm{M}^{j\rightarrow i}$.
Similar to the foreground point grouping process, we match $\bm{M}^i$ and $\bm{M}^{j\rightarrow i}$ based on their clusters' centers, where the $q$-th point cluster of the $i$-th agent $\bm{m}^{i;q}$ and the $r$-th point cluster of the $j$-th agent $\bm{m}^{j\rightarrow i;r}$ belong to the same object if their centers' distance $\Vert\bm{C}^{i;q}-\bm{C}^{j\rightarrow i;r}\Vert_2$ is less than a predefined threshold $\epsilon_\text{agg}$.
After matching, we organize all point clusters as two disjoint sets $\bm{M}_\text{unique}$ and $\bm{M}_\text{share}$. 
The set $\bm{M}_\text{unique}$ comprises point clusters exclusively observed by a single agent, which do not need to be involved in the following aggregation process.
Differently, $\bm{M}_\text{share}$ is the set of tuples including point clusters belonging to the same object in the scene. 
We combine each tuple to form a novel point cluster that encompasses comprehensive low-level and high-level object information.
In detail, assume that $\bm{m}^{i;q}$ and $\bm{m}^{j\rightarrow i;r}$ belong to the $s$-th object in the scene, the aggregated point cluster $\ddot{\bm{m}}^s=(\ddot{\bm{P}}^s, \ddot{\bm{C}}^s, \ddot{\bm{F}}^s, \ddot{\bm{B}}^s)$ can be formulated as follows:
\begin{equation}
\ddot{\bm{P}}^s=\bm{P}^{i;q}\cup \bm{P}^{j\rightarrow i;r},
\end{equation}
\begin{equation}
\ddot{\bm{C}}^s=\frac{\bm{C}^{i;q}+\bm{C}^{j\rightarrow i;r}}{2},
\end{equation}
\begin{equation}
\ddot{\bm{F}}^s=\text{avgpool}(\bm{F}^{i;q}, \bm{F}^{j\rightarrow i;r}),
\end{equation}
\begin{equation}
  \ddot{\bm{B}}^s =
    \begin{cases}
      \bm{B}^{i;q}, & \text{if $\hat{\bm{c}}^{i;q}-\hat{\bm{c}}^{j\rightarrow i;r}\geq0$} \\
       \bm{B}^{j\rightarrow i;r}, & \text{otherwise}
    \end{cases}
    ,
\end{equation}
Since there are no convolution operations and unnecessary zero-padding, the computational complexity of aggregating point clusters in our PCA is only strongly related to the number of potential objects, which is more efficient for long-range collaboration than BEV-based aggregation methods.

We denote the intermediate representation of point clusters in the scene after aggregation as $\ddot{\bm{M}}=\{\ddot{\bm{m}}^s\}_{s=1}^{N_\text{object}}$, where $N_\text{object}=\Vert\bm{M}_\text{unique}\Vert+\Vert\bm{M}_\text{share}\Vert$ is the potential object number in the scene observed by all involved agents.
Different from BEV-based message aggregation methods, $\ddot{\bm{P}}^s$ contains complete low-level structure information, which can be utilized to enhance the precision of the proposal bounding box $\ddot{\bm{B}}$.
In detail, we apply an additional SIR module that includes $L_2$ layers on $\ddot{\bm{M}}$, which predicts the box residual $\Delta_\text{res}$ to its corresponding ground truth box.
For each point cluster $\ddot{\bm{m}}^s$, we generate its point feature $\ddot{\bm{F}}^s_\text{point}$ by concatenating its offsets from the cluster proposal $\ddot{\bm{B}}^s$ and the cluster feature $\ddot{\bm{F}}^s$.
This introduces the proposal boundary information to the SIR module, which can handle the size ambiguity problem to a certain extent~\cite{li2021lidar}.
We define the residual loss $\mathcal{L}_\text{res}$ as the L1 distance between $\Delta_\text{res}$ and the ground-truth residual $\hat{\Delta}_\text{res}$.
In addition, we define the soft classification label as $\min(1, \max(0, 2u-0.5))$ following previous works~\cite{shi2020pv,shi2020points}, where $u$ is the 3D Intersection-of-Union (IoU) between the predicted proposal and the ground truth.
We adopt cross-entroy loss as the IoU loss $\mathcal{L}_\text{iou}$.
Considering all losses in our framework (i.e., \ref{sec:point_cluster_extraction}, \ref{sec:point_cluster_distribution}, and \ref{sec:point_cluster_aggregation}), the total training loss $\mathcal{L}$ can be formulated as:
\begin{equation}
    \mathcal{L}=\mathcal{L}_\text{seg}+\mathcal{L}_\text{center}+\mathcal{L}_\text{reg}+\mathcal{L}_\text{cls}+\mathcal{L}_\text{res}+\mathcal{L}_\text{iou}.
\end{equation}

\subsection{Robustness}
\label{sec:robustness}

In a realistic communication setting, the presence of pose error and time latency is unavoidable and it leads to misalignment of point clouds that significantly affect the reliability of transferred information in collaborative perception.
Benefiting from the low-level object information in point clusters, we propose parameter-free approaches that can generalize to different noise settings.
For simplicity, we take the correction processes that happen between the $i$-th (ego) and $j$-th agents as an example.

\textbf{Pose Correction}.
To address this issue, we propose to align the clusters from different agents belonging to the same object.
After receiving the message from the $j$-th agent, we first align the coordinate space of $\bm{M}^j$ to the $i$-th agent with relative pose $\xi^{j\rightarrow i} = {(\xi^i)}^{-1}\circ\xi^j$, where $\circ$ means multiplying their homogeneous transformation matrices.
We denote $\bm{m}^{i;q}$ and $\bm{m}^{j\rightarrow i;r}$ as point clusters belonging to the $s$-th unique object in the scene after spatial cluster matching with threshold $\epsilon_\text{pose}$.
In the following, we simplify each pose in 2D space.
The pose of the $s$-th unique object is defined as $\chi^s=\xi^i\circ(\bm{C}^{i;q}+\bm{C}^{j\rightarrow i;r})/2$.
Inspired by CoAlign~\cite{lu2023robust}, we define the pose consistency error vector as $\bm{e}^{js}=\bm{C}^{j;r}\circ({(\xi^j)}^{-1}\circ\chi^s)$, which is zero when there is no pose error.
The overall optimization problem can be formatted as follows:
\begin{equation}
    \{{(\chi^s)}^\prime,{(\xi^j)}^\prime\}=\mathop{\arg}_{\{\xi^j,\chi^s\}}\min\sum_{j=1}^{N_\text{agent}}\sum_{s=1}^{N_\text{object}}{(\bm{e}^{js})}^T\bm{e}^{js}.
\end{equation}
Different from existing BEV-based methods that need additional detection results for pose correction, we can directly utilize low-level information contained in the point cluster.

\textbf{Latency Compensation}.
SyncNet~\cite{lei2022latency} proposes to complete BEV maps at the current timestamp using locally stored historical BEV maps from other agents.
Differently, we propose to directly predict the location of point clusters in the current timestamp via speed estimation based on the low-level coordinate information, which is more efficient and interpretable.
We denote the received point clusters from the $j$-th agent at the $t$-th timestamp as $\bm{M}^{j;t-\tau^{j\rightarrow i;t}}$, where $\tau^{j\rightarrow i;t}$ is the time latency.
The stored point clusters from the $j$-th agent during the last communication round are denoted as $\bm{M}^{j;t^\prime}$, where $t^\prime < t-\tau^{j\rightarrow i;t}$.
Similar to the spatial matching process in \ref{sec:point_cluster_aggregation}, we match point clusters along the time dimension by measuring whether their Euclidean distance is in a predefined range $[\underline{\epsilon}_\text{latency},\overline{\epsilon}_\text{latency}]$.
Assume $\bm{m}^{j;r;t^\prime}$ and $\bm{m}^{j;q;t-\tau^{j\rightarrow i;t}}$ belong to the same object, we can infer its speed $\bm{v}^{j;q;t-\tau^{j\rightarrow i;t}}$ and offset $\Delta \bm{d}_\text{cluster}^{j;q;t-\tau^{j\rightarrow i;t}}$ to the current timestamp $t$ as follows: 
\begin{equation}
\bm{v}^{j;q;t-\tau^{j\rightarrow i;t}}=\frac{\Vert\bm{C}^{j;q;t-\tau^{j\rightarrow i;t}}-\bm{C}^{j;r;t^\prime}\Vert_2}{t-\tau^{j\rightarrow i;t}-t^\prime},
\end{equation}
\begin{equation}
\Delta \bm{d}_\text{cluster}^{j;q;t-\tau^{j\rightarrow i;t}}=\bm{v}^{j;q;t-\tau^{j\rightarrow i;t}}\times \tau^{j\rightarrow i;t}.
\end{equation}
Finally, we can obtain the estimated point cluster $\bm{m}^{j;q;t}$ at the $t$-th timestamp by coordinates translation with $\Delta \bm{d}_\text{cluster}^{j;q;t-\tau^{j\rightarrow i;t}}$.

\begin{figure}[t]
\centering
\includegraphics[width=\linewidth]{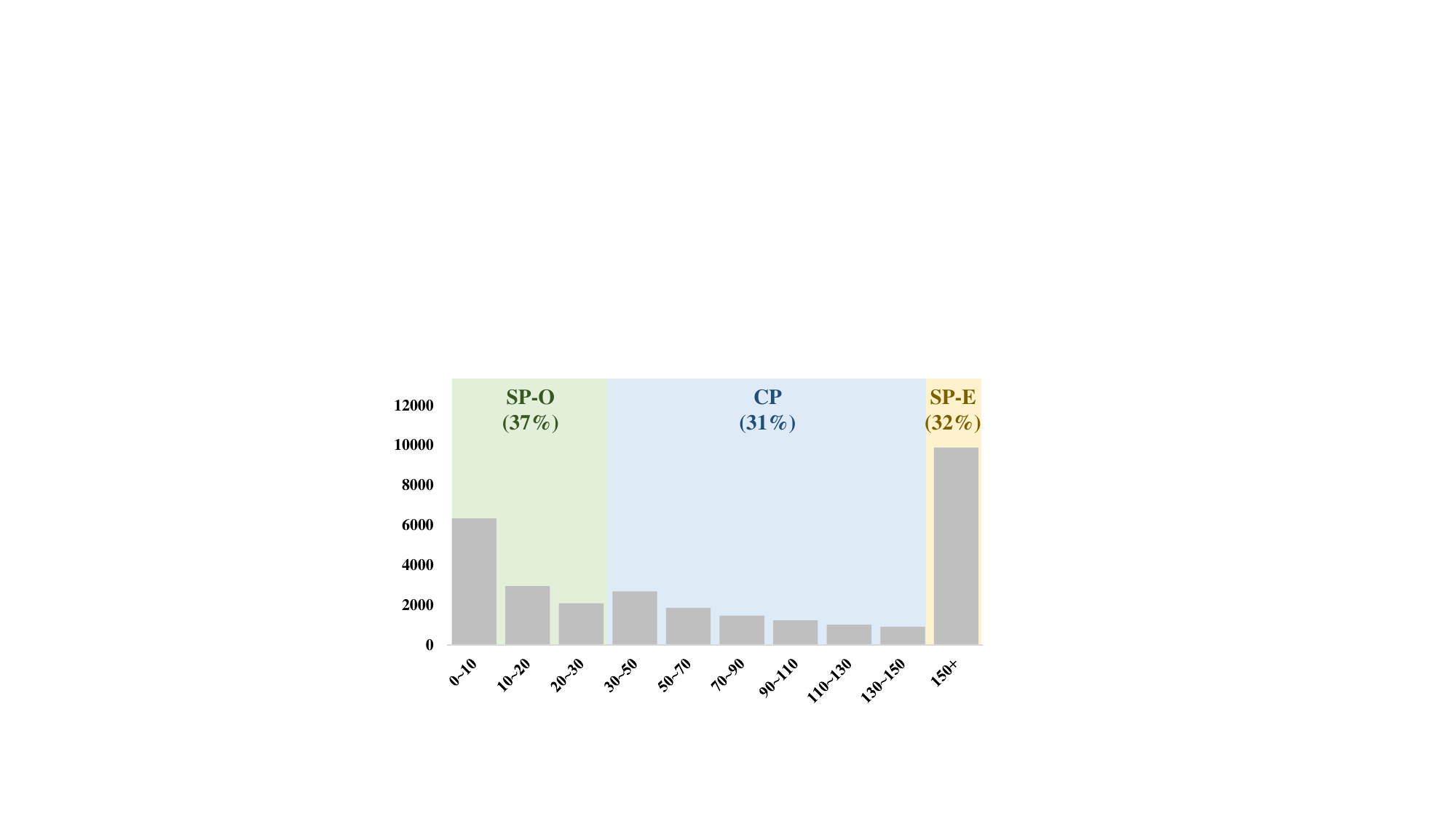}
\caption{Illustration of histogram of all targets and proportions of those belonging to SP-O, SP-E, and CP categories in the test set of DAIR-V2X-C.}
\label{fig:exp:new_metric}
\end{figure}

\section{Experiments}

\subsection{Datasets and Evaluation Metrics}

We conducted experiments on two widely used benchmarks for collaborative perception, focusing on the problem of collaborative 3D object detection.
The following section presents the details of our experiments:

\textbf{DAIR-V2X-C}~\cite{yu2022dair}. This is the first to provide a large-scale collection of real-world scenarios for vehicle-infrastructure collaborative autonomous driving.
It contains 38,845 frames of point cloud data annotated with almost 464k 3D bounding boxes representing objects in 10 different classes.
Since the original DAIR-V2X-C does not include objects beyond the camera's view, we have adopted the complemented annotations encompassing the 360-degree detection range, which are relabeled by Hu \textit{et al}.~\cite{hu2022where2comm}.

\textbf{V2XSet}~\cite{xu2022v2x}. This is a large-scale V2X perception dataset founded on CARLA~\cite{dosovitskiy2017carla} and OpenCDA~\cite{xu2021opencda}, which explicitly takes into account real-world noises like localization error and transmission latency. 
V2XSet has 11,447 frames (6,694/ 1,920/2,833 for train/validation/test respectively) captured in 55 representative simulation scenes that cover the most common driving scenarios in real life. 
Each scene typically involves 2-7 agents engaged in collaborative perception.

We follow previous works~\cite{hu2022where2comm, xu2022v2x} to select one of the agents in the scene as the ego agent, whose detection results are assessed by Average Precision (AP) at Intersection-over-Union (IoU) thresholds of 0.5 and 0.7, denoted as AP@0.5 and AP@0.7. 
Nevertheless, they treat all elements in the scenario uniformly, disregarding the input of collaborative participants.
To compare the collaborative perception ability of different methods in a more fine-grained manner, we calculate the number of points observed by the ego agent $N_\text{ego}$ in all target objects and categorize them as \textbf{S}ingle-agent \textbf{P}erception of \textbf{O}ther agents (SP-O), \textbf{S}ingle-agent \textbf{P}erception of \textbf{E}go agent (SP-E), and \textbf{C}ollaborative \textbf{P}erception (CP).
The proportions of all categories in the test set of DAIR-V2X-C are shown in Figure~\ref{fig:exp:new_metric}.
SP-O signifies objects that are scarcely perceived by the ego agent, while AP$_\text{SP-O}$ assesses the effectiveness of message packing in preserving full object details within bandwidth limitations.
CP indicates objects that are partially observable by the individual agent and require assistance from other agents to obtain comprehensive information about the objects.
AP$_\text{CP}$ serves as a metric to evaluate the effectiveness of the entire collaborative pipeline.
SP-E denotes objects that are more effectively scanned by the individual agent, with additional information provided by other agents primarily serving as supplementary.

\subsection{Implementation Details}
We set the perception range along the $x$, $y$, and $z$-axis to $[-140.8m, 140.8m]\times [-40m, 40m]\times [-3m, 1m]$  for V2XSet and $[-100.8m, 100.8m]\times [-40m, 40m]\times [-3m, 1m]$ for DAIR-V2X-C, respectively.
The communication results measure message size in bytes using a logarithmic scale with base 2.
The thresholds $\epsilon_\text{agg}$, $\epsilon_\text{pose}$, $\underline{\epsilon}_\text{latency}$, and $\overline{\epsilon}_\text{latency}$ for cluster matching are set as 0.6, 1.5, 0.5 and 2.0, respectively.
The number of SIR layers is $L_1=6$ in PCE and $L_2=3$ during message decoding.
The channel number of cluster features is $D=128$.
Adam~\cite{kingma2014adam} is employed as the optimizer for training our model end-to-end on NVIDIA Tesla V100 GPUs, with a total of 35 epochs.
The initial learning rate is set as 0.001 and we reduce it by 10 after 20 and 30 epochs, respectively.
Our method is implemented with PyTorch.

\begin{table}
    \centering
    \caption{Comparison with state-of-the-art methods on the test sets of V2XSet and DAIR-V2X-C on perfect setting.}
    \resizebox{\linewidth}{!}{
        \begin{tabular}{r||c|c|c|c}
        \hline\thickhline
        \rowcolor{mygray}
         & \multicolumn{2}{c|}{V2XSet} & \multicolumn{2}{c}{DAIR-V2X-C} \\
         \rowcolor{mygray}
         \multirow{-2}*{Method} & \multicolumn{1}{c}{AP@0.5} & \multicolumn{1}{c|}{AP@0.7} & \multicolumn{1}{c}{AP@0.5} & \multicolumn{1}{c}{AP@0.7} \\ \hline\hline
         DiscoNet~\cite{li2021learning} & 90.78 & 83.81 & 69.28 & 58.56 \\
         V2X-ViT~\cite{xu2022v2x} & 89.03 & 79.02 & 73.98 & 61.50 \\
        Where2comm~\cite{hu2022where2comm} & 85.03 & 76.77 & 71.48 & 60.36 \\
         OPV2V~\cite{xu2022opv2v} & 91.88 & 84.75 & 66.07 & 50.92 \\
         CoBEVT~\cite{xu2023cobevt} & 90.33 & 82.69 & 63.90 & 51.67 \\
         CoAlign~\cite{xu2022v2x} & -&- & 74.60 & 60.40 \\ \hline\hline
         Ours & \textbf{92.83} & \textbf{89.55} & \textbf{76.89} & \textbf{69.39} \\ \hline
        \end{tabular}
    }
    
    \label{tab:exp:sota}
\end{table}

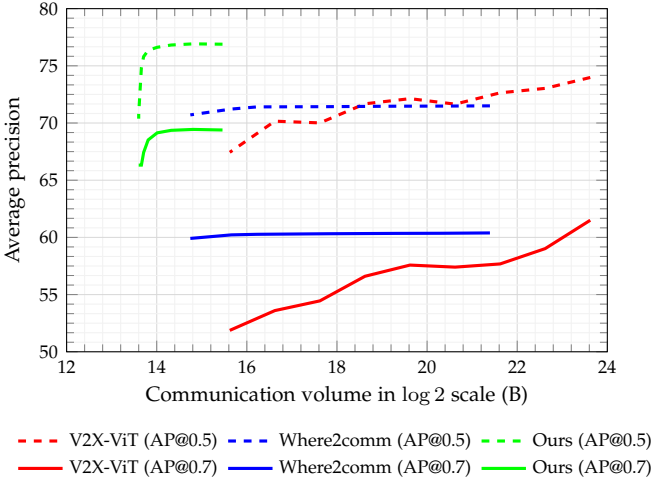
\begin{figure}[!t]
\centering
\resizebox{\linewidth}{!}{%
  \begin{tikzpicture}[/pgfplots/width=\linewidth, /pgfplots/height=0.7\linewidth]
    \begin{axis}[% Axis labels
                 ymin=50,ymax=80,xmin=12,xmax=24,
    			 % Axis labels
        		 xlabel=Communication volume in $\log 2$ scale (B),
        		 ylabel=Average precision,
         	   xlabel shift={-2pt},
        		 ylabel shift={-1pt},
                  label style={font=\footnotesize},
         		 % General appearance
		         font=\scriptsize,
		         % axis equal image=true,
		         enlargelimits=false,
		         clip=true,
		         % Grids
        	     grid style=solid, grid=both,
                  major grid style={white!85!black},
        		 minor grid style={white!95!black},
		 	 xtick={12,14,...,26},
                 xticklabels={12,14,16,18,20,22,24,26},
        		 ytick={50,55,...,80},
                 yticklabels={50,55,60,65,70,75,80},
                 minor x tick num=4,
                 minor y tick num=5,
                 legend columns=3,
        		 % Legend
        		 legend style={
                    at={(0.5,-0.3)},
                	anchor=center,
                    draw=none,
                    /tikz/every even column/.append style={column sep=0.05cm}
                 },
                 % legend cell align={left}
            ]
    \addplot+[red,dashed,mark=none,very thick] table[x=cv_v2xvit,y=v2xvit_50]{figs/figure1_final.txt};
    \addlegendentry{V2X-ViT (AP@0.5)}
    \addplot+[blue,dashed,mark=none,very thick] table[x=cv_where2comm,y=where2comm_50]{figs/figure1_final.txt};
    \addlegendentry{Where2comm (AP@0.5)}
    \addplot+[green,dashed,mark=none,very thick] table[x=cv_ours,y=ours_50]{figs/figure1_final.txt};
    \addlegendentry{Ours (AP@0.5)}
    \addplot+[red,solid,mark=none,very thick] table[x=cv_v2xvit,y=v2xvit_70]{figs/figure1_final.txt};
    \addlegendentry{V2X-ViT (AP@0.7)}
    \addplot+[blue,solid,mark=none,very thick] table[x=cv_where2comm,y=where2comm_70]{figs/figure1_final.txt};
    \addlegendentry{Where2comm (AP@0.7)}
    \addplot+[green,solid,mark=none,very thick] table[x=cv_ours,y=ours_70]{figs/figure1_final.txt};
    \addlegendentry{Ours (AP@0.7)}
    \end{axis}
\end{tikzpicture}}  
\caption{Comparison with state-of-the-art methods on the test sets of DAIR-V2X-C considering the performance-bandwidth trade-off.}
\label{fig:exp:bandwidth}
\end{figure}

\begin{figure}[!t]
\centering
% \begin{subfigure}[!t]{1\linewidth}
    \subfloat[Positional Error]{
    \centering
    \resizebox{\linewidth}{!}{%
      \begin{tikzpicture}[/pgfplots/width=\linewidth, /pgfplots/height=0.7\linewidth]
        \begin{axis}[% Axis labels
                     ymin=30,ymax=95,xmin=0,xmax=0.6,
        			 % Axis labels
            		 xlabel=Std ($m$),
            		 ylabel=Average Precision,
             	   xlabel shift={-2pt},
            		 ylabel shift={-1pt},
                      label style={font=\footnotesize},
             		 % General appearance
    		         font=\scriptsize,
    		         % axis equal image=true,
    		         enlargelimits=false,
    		         clip=true,
    		         % Grids
            	     grid style=solid, grid=both,
                      major grid style={white!85!black},
            		 minor grid style={white!95!black},
    		 	 xtick={0,0.1,...,0.6},
                     xticklabels={0,0.1,0.2,0.3,0.4,0.5,0.6},
            		 ytick={30,40,...,90},
                     yticklabels={30,40,50,60,70,80,90},
                     minor x tick num=4,
                     minor y tick num=5,
                     legend columns=3,
            		 % Legend
            		 legend style={
                        at={(0.5,-0.3)},
                    	anchor=center,
                        draw=none,
                        /tikz/every even column/.append style={column sep=0.05cm}
                     },
                     % legend cell align={left}
                ]
        \addplot+[red,dashed,mark=none,very thick] table[x=error,y=v2xvit_50]{figs/figure2_final.txt};
        \addlegendentry{V2X-ViT (AP@0.5)}
        \addplot+[blue,dashed,mark=none,very thick] table[x=error,y=where2comm_50]{figs/figure2_final.txt};
        \addlegendentry{Where2comm (AP@0.5)}
        \addplot+[green,dashed,mark=none,very thick] table[x=error,y=ours_50]{figs/figure2_final.txt};
        \addlegendentry{Ours (AP@0.5)}
        \addplot+[red,solid,mark=none,very thick] table[x=error,y=v2xvit_70]{figs/figure2_final.txt};
        \addlegendentry{V2X-ViT (AP@0.7)}
        \addplot+[blue,solid,mark=none,very thick] table[x=error,y=where2comm_70]{figs/figure2_final.txt};
        \addlegendentry{Where2comm (AP@0.7)}
        \addplot+[green,solid,mark=none,very thick] table[x=error,y=ours_70]{figs/figure2_final.txt};
        \addlegendentry{Ours (AP@0.7)}
        \end{axis}
    \end{tikzpicture}}  
    \label{fig:positional_error}}\\
    
    \subfloat[Heading Error]{
    \centering
    \resizebox{\linewidth}{!}{%
      \begin{tikzpicture}[/pgfplots/width=\linewidth, /pgfplots/height=0.7\linewidth]
        \begin{axis}[% Axis labels
                     ymin=30,ymax=95,xmin=0,xmax=0.6,
        			 % Axis labels
            		 xlabel=Std ($^\circ$),
            		 ylabel=Average Precision,
             	   xlabel shift={-2pt},
            		 ylabel shift={-1pt},
                      label style={font=\footnotesize},
             		 % General appearance
    		         font=\scriptsize,
    		         % axis equal image=true,
    		         enlargelimits=false,
    		         clip=true,
    		         % Grids
            	     grid style=solid, grid=both,
                      major grid style={white!85!black},
            		 minor grid style={white!95!black},
    		 	 xtick={0,0.1,...,0.6},
                     xticklabels={0,0.1,0.2,0.3,0.4,0.5,0.6},
            		 ytick={30,40,...,90},
                     yticklabels={30,40,50,60,70,80,90},
                     minor x tick num=4,
                     minor y tick num=5,
                     legend columns=3,
            		 % Legend
            		 legend style={
                        at={(0.5,-0.3)},
                    	anchor=center,
                        draw=none,
                        /tikz/every even column/.append style={column sep=0.05cm}
                     },
                     % legend cell align={left}
                ]
        \addplot+[red,dashed,mark=none,very thick] table[x=error,y=v2xvit_50]{figs/figure3_final.txt};
        \addlegendentry{V2X-ViT (AP@0.5)}
        \addplot+[blue,dashed,mark=none,very thick] table[x=error,y=where2comm_50]{figs/figure3_final.txt};
        \addlegendentry{Where2comm (AP@0.5)}
        \addplot+[green,dashed,mark=none,very thick] table[x=error,y=ours_50]{figs/figure3_final.txt};
        \addlegendentry{Ours (AP@0.5)}
        \addplot+[red,solid,mark=none,very thick] table[x=error,y=v2xvit_70]{figs/figure3_final.txt};
        \addlegendentry{V2X-ViT (AP@0.7)}
        \addplot+[blue,solid,mark=none,very thick] table[x=error,y=where2comm_70]{figs/figure3_final.txt};
        \addlegendentry{Where2comm (AP@0.7)}
        \addplot+[green,solid,mark=none,very thick] table[x=error,y=ours_70]{figs/figure3_final.txt};
        \addlegendentry{Ours (AP@0.7)}
        \end{axis}
    \end{tikzpicture}}  
    \label{fig:heading_error}}
\caption{Comparison with state-of-the-art methods on the test sets of V2XSet with pose noises following Gaussian distribution with standard deviations from $\{0.0, 0.1, 0.2, 0.3, 0.4, 0.5, 0.6\}$ for heading error ($^\circ$) and positional error ($m$), respectively.}
\label{fig:exp:noisy}
\end{figure}
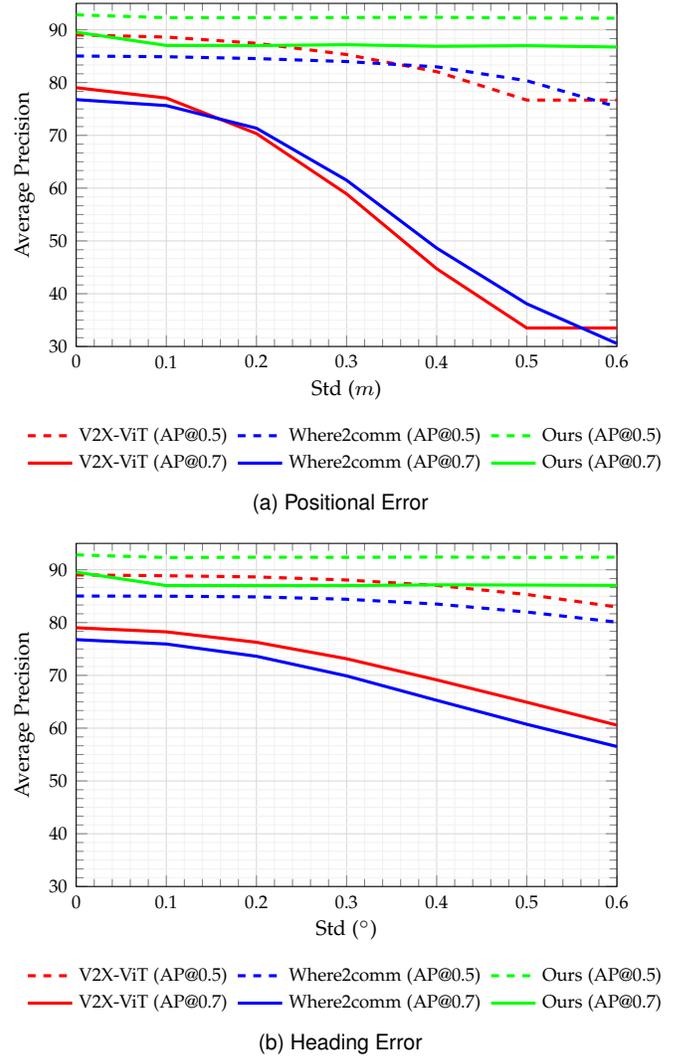

\subsection{Comparison With State-of-the-art Methods}
We conduct experiments using two collaborative perception datasets to compare our approach with previous state-of-the-art methods.
As shown in Table~\ref{tab:exp:sota}, our V2X-PC outperforms previous BEV-based approaches on the test sets of V2XSet and DAIR-V2X-C, indicating the effectiveness of adopting point cluster as the basic collaborative message unit in both simulated and realistic scenarios.
Our V2X-PC outperforms other methods with significant margins in the most stringent metric AP@0.7, i.e., 4.20\% and 8.99\% in V2XSet and DAIR-V2X-C respectively, demonstrating its ability to accurately locate the correct object through collaborative perception and regress a more precise bounding box to cover the object.

In real-world applications, collaborative perception methods must achieve a delicate equilibrium between communication volume and precision due to the typically limited and variable communication bandwidth. 
The communication volume is calculated as follows:
\begin{equation}
    \text{Comm}=\log_2(N\times C\times 16/8),
\end{equation}
where $N$ represents the number of collaborative message units, $C$ represents the number of channels, and the data is transmitted in fp16 data type, resulting in minimal performance impact.
The volume in bits is then converted to bytes using the logarithm base 2.
We explore the performance-bandwidth trade-off of our V2X-PC and previous typical methods in Figure~\ref{fig:exp:bandwidth}.
Thanks to the sparse nature of point clusters, the communication volume is under 16 even if we pack all cluster points, which is close to the lower bound of bandwidth usage of other methods. 
Under the same communication volume, our method achieves significant performance improvements of around 10.0 AP@0.7.
The little performance drops with bandwidth decrease indicate that appropriate point sampling strategies can reduce representation redundancy. 
\begin{figure}[!t]
\centering
\resizebox{\linewidth}{!}{%
  \begin{tikzpicture}[/pgfplots/width=\linewidth, /pgfplots/height=0.7\linewidth]
    \begin{axis}[% Axis labels
                 ymin=50,ymax=80,xmin=0,xmax=500,
    			 % Axis labels
        		 xlabel=Time latency ($ms$),
        		 ylabel=Average precision,
         	   xlabel shift={-2pt},
        		 ylabel shift={-1pt},
                  label style={font=\footnotesize},
         		 % General appearance
		         font=\scriptsize,
		         % axis equal image=true,
		         enlargelimits=false,
		         clip=true,
		         % Grids
        	     grid style=solid, grid=both,
                  major grid style={white!85!black},
        		 minor grid style={white!95!black},
		 	 xtick={0,100,...,500},
                 xticklabels={0,100,200,300,400,500},
        		 ytick={50,55,...,80},
                 yticklabels={50,55,60,65,70,75,80},
                 minor x tick num=4,
                 minor y tick num=5,
                 legend columns=3,
        		 % Legend
        		 legend style={
                    at={(0.5,-0.3)},
                	anchor=center,
                    draw=none,
                    /tikz/every even column/.append style={column sep=0.05cm}
                 },
                 % legend cell align={left}
            ]
    \addplot+[red,dashed,mark=none,very thick] table[x=latency,y=v2xvit_50]{figs/figure4_final.txt};
    \addlegendentry{V2X-ViT (AP@0.5)}
    \addplot+[blue,dashed,mark=none,very thick] table[x=latency,y=where2comm_50]{figs/figure4_final.txt};
    \addlegendentry{Where2comm (AP@0.5)}
    \addplot+[green,dashed,mark=none,very thick] table[x=latency,y=ours_50]{figs/figure4_final.txt};
    \addlegendentry{Ours (AP@0.5)}
    \addplot+[red,solid,mark=none,very thick] table[x=latency,y=v2xvit_70]{figs/figure4_final.txt};
    \addlegendentry{V2X-ViT (AP@0.7)}
    \addplot+[blue,solid,mark=none,very thick] table[x=latency,y=where2comm_70]{figs/figure4_final.txt};
    \addlegendentry{Where2comm (AP@0.7)}
    \addplot+[green,solid,mark=none,very thick] table[x=latency,y=ours_70]{figs/figure4_final.txt};
    \addlegendentry{Ours (AP@0.7)}
    \end{axis}
\end{tikzpicture}}
\caption{Comparison with state-of-the-art methods on the test sets of DAIR-V2X-C with time latency of $100ms$, $200ms$, $300ms$, $400ms$, and $500ms$.}
\label{fig:exp:latency}
\end{figure}
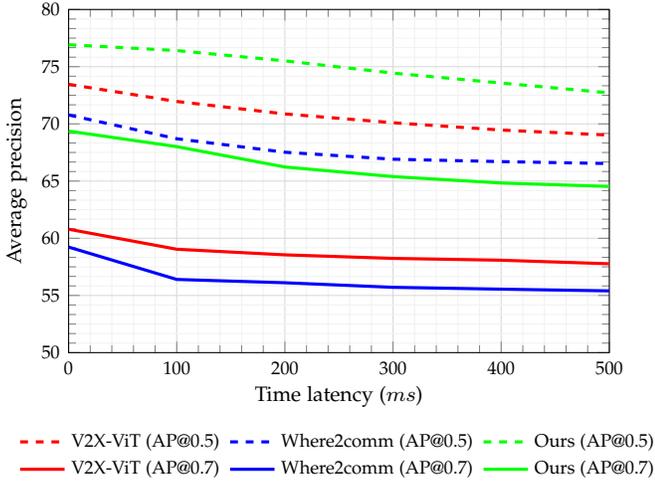
Collaborative agents depend on precise pose data from others to transform the coordinates of received messages. 
Despite advanced localization technologies like GPS, pose error is unavoidable. 
Therefore, collaborative approaches need to be resilient to localization errors.
In Figure~\ref{fig:exp:noisy}, we compare different methods with pose noises following Gaussian distribution with standard deviations from $\{0.0, 0.1, 0.2, 0.3, 0.4, 0.5, 0.6\}$ for heading error ($^\circ$) and positional error ($m$), respectively.
Results show that there is no significant AP drop, which validates that our V2X-PC can handle large noise disturbances without further finetuning and additional model parameters.

Time latency poses a pervasive challenge in real-world V2X communication, leading to asynchronization between ego features and received collaborative features from others. 
Since practical collaborative perception methods are required to exhibit robustness against time latency, we compare the model robustness against time latency ranging from 0 to 500$ms$ in Figure~\ref{fig:exp:latency}.
Compared to previous intermediate methods, our V2X-PC achieves superior performance, with similar AP degradation as latency increases.
It is worth noting that both V2X-ViT and Where2comm need to be finetuned with data on different noisy levels, while our V2X-PC can adapt to arbitrary noise levels in a zero-shot manner.

\begin{table}
    \centering
    \caption{Comparison with state-of-the-art methods on the test set of DAIR-V2X-C with different target categories.}
    \resizebox{0.8\linewidth}{!}{
        \begin{tabular}{r||c|c|c}
        \hline\thickhline
        \rowcolor{mygray}
         & \multicolumn{3}{c}{Metric} \\
         \rowcolor{mygray}
         \multirow{-2}*{Method} & \multicolumn{1}{c}{AP$_\text{SP-O}$} & \multicolumn{1}{c}{AP$_\text{CP}$} & \multicolumn{1}{c}{AP$_\text{SP-E}$} \\ \hline\hline
         Where2comm & 38.54 & 73.17 & 67.18  \\
         V2X-ViT & 36.85 & 74.94 & 67.27  \\
         Ours & \textbf{40.13} & \textbf{82.63} & \textbf{76.72} \\ \hline
        \end{tabular}
    }
    \label{tab:exp:new_metric}
\end{table}

Collaborative perception aims to enhance the performance of the single vehicle, but the existing evaluation metrics (AP@0.5/AP@0.7) treat all objects equally which is not conducive to fine-grained analysis. 
As shown in Table~\ref{tab:exp:new_metric}, we split targets into different categories based on the number of points scanned by the ego agent and evaluate AP@0.7 for each category, denoted as AP$_\text{SP-O}$, AP$_\text{CP}$, and AP$_\text{SP-E}$.
Experimental results demonstrate the superiority of our V2X-PC over the state-of-the-art BEV-based approaches across all evaluation metrics.
In detail, high AP$_\text{SP-O}$ means that our PCP module can keep more complete object information since we avoid feature destruction caused by channel compression and spatial selection.
The results in the 2-nd column demonstrate the significant improvement of our V2X-PC over BEV-based methods in terms of the AP$_\text{CP}$ metric. This indicates that utilizing point clusters as the basic collaborative message unit is advantageous for both message packing and aggregation phases, resulting in enhanced collaboration compared to earlier techniques relying on dense BEV maps.

\subsection{Ablation Studies}

We assess various designs of our framework through ablation studies using the V2XSet and DAIR-V2X-C datasets.

\begin{table}
    \centering
    \caption{Different numbers of feature channels in PCE.}
    \resizebox{\linewidth}{!}{
        \begin{tabular}{r||c|c|c|c|c|c}
        \hline\thickhline
        \rowcolor{mygray}
         & \multicolumn{6}{c}{Channel Number} \\
         \rowcolor{mygray}
         \multirow{-2}*{Metric} & \multicolumn{1}{c}{8} & \multicolumn{1}{c}{16} & \multicolumn{1}{c}{32} & \multicolumn{1}{c}{64} & \multicolumn{1}{c}{96} & \multicolumn{1}{c}{128} \\ \hline\hline
         AP@50 & 74.41 & 75.32 & 77.12 & 76.63 & 77.40 & 77.30 \\
         AP@70 & 64.79 & 67.08 & 69.32 & 68.92 & 69.55 & 69.43 \\\hline
        \end{tabular}
    }
    \label{tab:exp:feat_ch}
\end{table}
\textbf{Number of Feature Channels in PCE}.
We assess AP@50 and AP@70 using various channel numbers of cluster features in PCE on the test set of DAIR-V2X-C in Table~\ref{tab:exp:feat_ch}.
The results indicate that there is no notable decrease in performance when the number of channels is reduced to 16.
We argued that since we explicitly include structure representation in point clusters, the semantic information can be compressed to a large extent, leading to small bandwidth consumption.
In contrast to prior approaches that compress solely before the message packing phase, our framework incorporates a small number of channels throughout the entire encoding phase, thereby decreasing network size and computational overhead without destroying object features across channels ($3$-rd and $4$-st rows).

\begin{table}
    \centering
    \caption{Sampling cluster points with different ratios and methods.}
    \resizebox{\linewidth}{!}{
        \begin{tabular}{r||c|c|c|c|c|c}
        \hline\thickhline
        \rowcolor{mygray}
         & \multicolumn{6}{c}{Ratio} \\
         \rowcolor{mygray}
         \multirow{-2}*{Method} & \multicolumn{1}{c}{$1/128$} & \multicolumn{1}{c}{$1/64$} & \multicolumn{1}{c}{$1/32$} & \multicolumn{1}{c}{$1/16$} & \multicolumn{1}{c}{$1/8$} & \multicolumn{1}{c}{$1/4$} \\ \hline\hline
         RPS & 64.60 & 65.46 & 67.12 & 67.80 & 68.61 & 68.75 \\ 
         FPS & 65.39 & 66.32 & 67.43 & 68.52 & 69.14 & 69.36\\ 
         S-FPS & 66.21 & 67.08 & 68.04 & 69.01 & 69.05 & 68.96 \\ 
         D-FPS & 65.65 & 66.52 & 67.51 & 68.57 & 69.21 & 69.46 \\ 
         SD-FPS & 66.12 & 67.03 & 68.08 & 69.22 & 69.19 & 69.41 \\ \hline
        \end{tabular}
    }
    \label{tab:exp:sam_rat}
\end{table}
\textbf{Sampling Point Clusters with Different Ratios and Methods}.
We evaluate AP@70 of different sampling ratios and methods during message packing on the test set of DAIR-V2X-C in Table~\ref{tab:exp:sam_rat}.
The ratio is the number of sample points divided by the total number of points.
The first and second rows represent the baseline methods of random sampling and basic FPS. The second, third, and fourth rows represent the results of introducing the semantic score, density score, and their joint application. It can be seen that the semantic score performs better than the baseline under smaller sampling ratios, because it preserves the semantic of object categories in extremely sparse structure. The density score performs better than the baseline under larger sampling ratios, because it can remove local redundant information. By combining the two scores, our method has significant performance advantages over the baseline at all sampling ratios.

\begin{table}
    \centering
    \caption{Features for object pose calculation during pose correction. ``0.1'' label of each column means experiments with standard deviation 0.1 for both heading error ($^\circ$) and positional error ($m$).}
    \resizebox{\linewidth}{!}{
        \begin{tabular}{r||c|c|c|c|c}
        \hline\thickhline
        \rowcolor{mygray}
         & \multicolumn{5}{c}{Pose Error} \\
         \rowcolor{mygray}
         \multirow{-2}*{Feature} & \multicolumn{1}{c}{0.1} & \multicolumn{1}{c}{0.2} & \multicolumn{1}{c}{0.3} & \multicolumn{1}{c}{0.4} & \multicolumn{1}{c}{0.5}  \\ \hline\hline
         Point Center & 75.13/45.44 & 76.89/46.59 & 78.50/48.20 & 79.60/51.03 & 78.90/52.07 \\ 
         Cluster Center & 90.22/86.95 & 90.34/87.01 & 90.04/87.06 & 90.10/87.06 & 89.97/86.94 \\ \hline
        \end{tabular}
    }
    \label{tab:exp:obj_pose}
\end{table}
\begin{table}
    \centering
    \caption{Lower bound for matching during latency compensation.}
    \resizebox{\linewidth}{!}{
        \begin{tabular}{r||c|c|c|c|c}
        \hline\thickhline
        \rowcolor{mygray}
         & \multicolumn{5}{c}{Time Latency} \\
         \rowcolor{mygray}
         \multirow{-2}*{$\underline{\epsilon}_\text{latency}$} & \multicolumn{1}{c}{100} & \multicolumn{1}{c}{200} & \multicolumn{1}{c}{300} & \multicolumn{1}{c}{400} & \multicolumn{1}{c}{500}  \\ \hline\hline
         0 & 76.36/67.97 & 75.23/66.08 & 74.13/64.61 & 73.19/63.57 & 71.84/62.79 \\ 
         0.1 & 76.34/67.93 & 75.20/66.09 &  74.15/64.60 & 73.20/63.53 & 71.82/62.80 \\ 
         0.2 & 76.34/67.88 & 75.21/66.06 & 74.13/64.59 & 73.17/63.46 & 72.37/64.00 \\ 
         0.3 & 76.34/67.86 & 75.19/66.00 & 74.14/64.60 & 73.19/63.50 & 72.72/64.46 \\ 
         0.4 & 76.42/68.01 & 75.26/66.05 & 74.11/64.58 & 73.50/64.42 & 72.76/64.47 \\ 
         0.5 & 76.44/68.01 & 75.35/65.54 & 74.36/65.18 & 73.79/64.82 & 72.75/64.48 \\ \hline
        \end{tabular}
    }
    \label{tab:exp:low_bound}
\end{table}
\textbf{Features for Object Pose Calculation during Pose Correction}.
We evaluate AP@0.5/AP@0.7 of different features for object pose calculation during pose correction in Table~\ref{tab:exp:obj_pose}.
``Point Center'' denotes representing the object pose with the mean coordinates of cluster points.
``Cluster Center'' denotes representing the object pose with the estimated cluster center.
The experiments demonstrate that utilizing the ``Point Center'' method for determining the object's pose correction results in notable performance deterioration.
Due to Lidar typically scanning objects partially, directly representing a point cluster by averaging the coordinates of all cluster points can lead to a significant offset from the true object center.

\textbf{Lower Bound for Matching during Latency Compensation}.
We evaluate AP@0.5/AP@0.7 of different lower bounds $\underline{\epsilon}_\text{latency}$ for matching during latency compensation in Table~\ref{tab:exp:low_bound}.
In line with Figure~\ref{fig:exp:latency}, an increase in time latency may result in performance degradation due to notable position shifts that complicate temporal alignment.
If our method encounters relatively high time latency (e.g., 300$ms$, 400$ms$, and 500$ms$), performance significantly decreases when $\underline{\epsilon}_\text{latency}=0$ or is very low.
Our research revealed that certain vehicles remain stationary, rendering the assumption of uniform motion in Section~\ref{sec:robustness} invalid.
By adjusting $\underline{\epsilon}_\text{latency}$ as demonstrated in the 3-rd, 4-th, and 5-th columns, we can mitigate this issue by filtering out stationary targets during latency compensation.

\subsection{Qualitative Analysis}

\begin{figure}[!t]
\centering
\includegraphics[width=\linewidth]{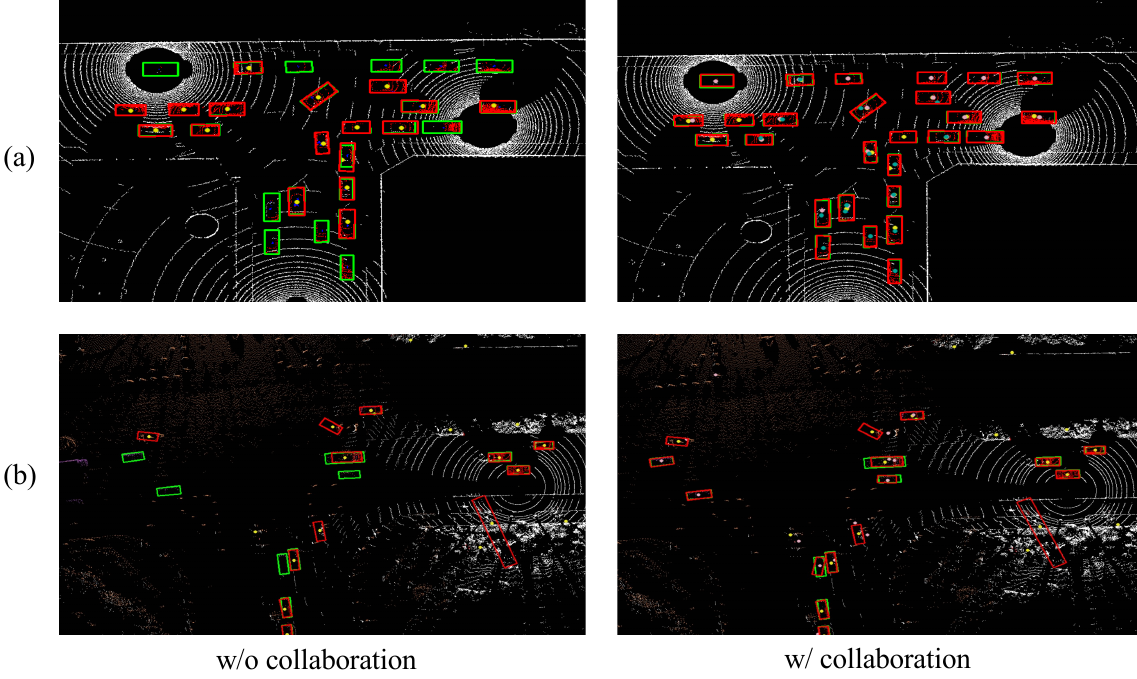}
\caption{Qualitative comparison results of our V2X-PC with and without collaboration on the test sets of (a) V2XSet and (b) DAIR-V2X-C, respectively. The green bounding boxes represent the ground-truth, and the red ones depict our predictions consistently throughout the subsequent qualitative results.}
\label{fig:exp:coop}
\end{figure}

\begin{figure}[!t]
\centering
\includegraphics[width=\linewidth]{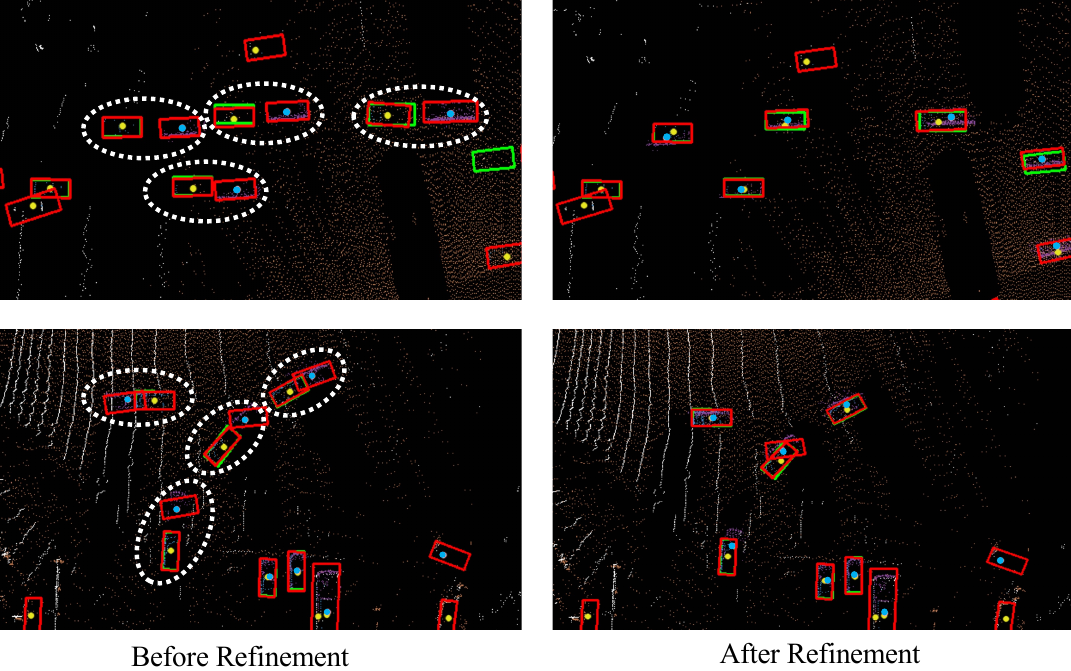}
\caption{Qualitative comparison results of our V2X-PC before and after latency compensation on the test set of DAIR-V2X-C dataset. The yellow and blue dots in the bounding boxes denotes cluster centers from the ego car and the road infrastructure, respectively. The false-positive results are highlighted by white dash ovals.}
\label{fig:exp:time_latency}
\end{figure}

As shown in Figure~\ref{fig:exp:coop}, we show qualitative evaluation of our V2X-PC w/ and w/o collaboration on the test sets of both V2XSet and DAIR-V2X-C datasets.
We can observe that the absence of collaboration leads to the overlooking of many objects to the ego agent's inability to perceive sufficient information regarding long-range or obscured targets.
These issue can be effectively addressed using our message packing and aggregation mechanisms that are based on point clusters.
Af

\begin{figure}[!t]
\centering
\includegraphics[width=\linewidth]{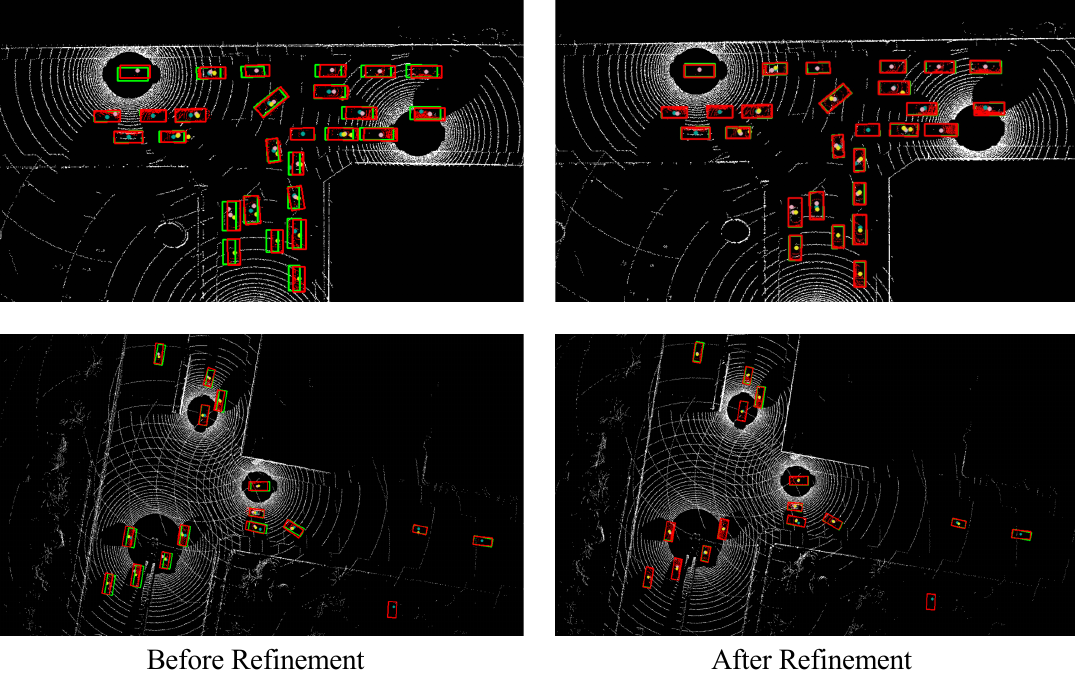}
\caption{Qualitative comparison results before and after pose correction on the test set of V2XSet dataset.}
\label{fig:exp:pose}
\end{figure}

We also visualize the detection results before and after latency compensation on the test set of the DAIR-V2X-C dataset in Figure~\ref{fig:exp:time_latency}. 
The latency results in a delay of point clusters from road infrastructure (marked by yellow dots) compared to the point clusters from the ego car (marked by blue dots) at the current timestamp. 
Consequently, the ego car erroneously detects them as distinct objects, resulting in an escalation of false-positive predictions.
After refinement by our latency compensation module, the delayed point clusters can be adjusted to the correct positions, enhancing the detection results.
Further, we illustrate the detection results before and after pose correction on the test set of the V2XSet dataset in Figure~\ref{fig:exp:pose}.
By aligning clusters among all agents, we can correct the noisy poses and obtain bounding boxes with high precision.

\section{Conclusion and Discussion}
In this paper, we focus on the vehicle-to-everything collaborative perception task, which improves the single-vehicle perception capability by integrating complementary information from surrounding traffic agents.
Three main scientific problems are unveiled for the collaborative perception task, encompassing bandwidth-friendly message packing, efficient message aggregation, and structure representation communication.
In this context, we argue that existing BEV-based methods suffer from object feature destruction during message packing, inefficient message aggregation for long-range collaboration, and implicit structure representation communication.
In pursuit of successfully handling the aforementioned issues, we create a brand new message unit for collaborative perception, namely point cluster, and based on this we further present a novel collaborative framework V2X-PC.
Regarding message packing, we propose a Point Cluster Packing (PCP) module to decrease the number of cluster points while preserving the geometric structure and preventing high-level object information loss.
To achieve effective message aggregation with high efficiency, we design a Point Cluster Aggregation (PCA) module to match and merge point clusters from different sources belonging to the same object.
Without the dependency on square BEV maps and convolution operations, our PCA is range-irrelevant and more efficient for long-range communication.
Moreover, the geometric structural information maintained in the point clusters supports explicit structure representation communication, which can improve the precision of predictions.
To deal with time latency and pose errors encountered in real environments, we align point clusters from spatial and temporal dimensions and propose parameter-free solutions for them.
Extensive experiments on two collaborative perception benchmarks show our method outperforms previous state-of-the-art methods.

\textbf{Limitation and Future Work}. By leveraging the low-level information within point clusters, we introduce parameter-free solutions to enhance the robustness of our approach.
Despite good results, we may need to manually adjust hyperparameters when facing a new environment. 
In the future, we plan to address this issue using novel techniques specifically designed for adapting to environmental changes.
In addition, we also plan to extend our V2X-PC to handle tasks considering temporal modeling, like tracking and forecasting~\cite{yu2023v2x}.

% if have a single appendix:
%\appendix[Proof of the Zonklar Equations]
% or
%\appendix  % for no appendix heading
% do not use \section anymore after \appendix, only \section*
% is possibly needed

% use appendices with more than one appendix
% then use \section to start each appendix
% you must declare a \section before using any
% \subsection or using \label (\appendices by itself
% starts a section numbered zero.)
%

% \appendices
% \section{Proof of the First Zonklar Equation}
% Appendix one text goes here.

% you can choose not to have a title for an appendix
% if you want by leaving the argument blank
% \section{}
% Appendix two text goes here.

% % use section* for acknowledgment
% \ifCLASSOPTIONcompsoc
%   % The Computer Society usually uses the plural form
%   \section*{Acknowledgments}
% \else
%   % regular IEEE prefers the singular form
%   \section*{Acknowledgment}
% \fi

% The authors would like to thank...

% Can use something like this to put references on a page
% by themselves when using endfloat and the captionsoff option.
\ifCLASSOPTIONcaptionsoff
  \newpage
\fi

% trigger a \newpage just before the given reference
% number - used to balance the columns on the last page
% adjust value as needed - may need to be readjusted if
% the document is modified later
%\IEEEtriggeratref{8}
% The "triggered" command can be changed if desired:
%\IEEEtriggercmd{\enlargethispage{-5in}}

% references section

% can use a bibliography generated by BibTeX as a .bbl file
% BibTeX documentation can be easily obtained at:
% http://mirror.ctan.org/biblio/bibtex/contrib/doc/
% The IEEEtran BibTeX style support page is at:
% http://www.michaelshell.org/tex/ieeetran/bibtex/
\bibliographystyle{IEEEtran}
% argument is your BibTeX string definitions and bibliography database(s)
\bibliography{main}

% that's all folks
\end{document}